\documentclass{article}
\PassOptionsToPackage{numbers, compress}{natbib}
\usepackage[final]{nips_2016}
\usepackage[utf8]{inputenc} 
\usepackage[T1]{fontenc}    
\usepackage{url}            
\usepackage{booktabs}       
\usepackage{amsfonts}       
\usepackage{nicefrac}       
\usepackage{microtype}      
\usepackage{algorithm}
\usepackage{algpseudocode}
\usepackage{amsmath}
\usepackage{graphicx}
\usepackage{subfig}

\usepackage{etoolbox}

\newsavebox{\mybox}
\newenvironment{mytabularwrap}{\begin{lrbox}{\mybox}}
	{\end{lrbox}%
	\setbox0\hbox{\usebox\mybox}%
	\ifdim\wd0<\textwidth
	\usebox\mybox%
	\else
	\resizebox{\textwidth}{!}{\usebox\mybox}%
	\fi
}

\BeforeBeginEnvironment{tabular}{\begin{mytabularwrap}}
	\AfterEndEnvironment{tabular}{\end{mytabularwrap}}

\title{Similarity Mapping with Enhanced Siamese Network \\for Multi-Object Tracking}
\author{
	Minyoung~Kim\\
	Panasonic Silicon Valley Laboratory\\
	Cupertino, CA\\
	\texttt{minyoung.kim@us.panasonoic.com} \\
	\And
	Stefano~Alletto \\
	University of Modena and Reggio Emilia \\
	Modena, MO \\
	\texttt{stefano.alletto@unimore.it} \\
	\And
	Luca Rigazio \\
	Panasonic Silicon Valley Laboratory\\
	Cupertino, CA\\
	\texttt{luca.rigazio@us.panasonic.com} \\
}

\begin{document}
	
\maketitle	
	
\begin{abstract}
  Multi-object tracking has recently become an important area of computer vision, especially for Advanced Driver Assistance Systems (ADAS). Despite growing attention, achieving high performance tracking is still challenging, with state-of-the-art systems resulting in high complexity with a large number of hyper parameters. In this paper, we focus on reducing overall system complexity and the number hyper parameters that need to be tuned to a specific environment. We introduce a novel tracking system based on similarity mapping by Enhanced Siamese Neural Network (ESNN), which accounts for both appearance and geometric information, and is trainable end-to-end. Our system achieves competitive performance in both speed and accuracy on MOT16 challenge and KITTI benchmarks, compared to known state-of-the-art methods.
\end{abstract} 

\section{Introduction} \label{sec:Introduction}
Object tracking has been evolving rapidly, becoming a very active area of research in machine vision. Several approaches have been proposed to improve tracking performance \cite{DBLP:journals/corr/abs-1303-4803}, with various applications from surveillance systems \cite{Hu:2004:SVS:2220414.2220805} to autonomous driving \cite{Geiger2012CVPR}, and even  sports analytics \cite{WuLimYang13}. One major limitation of object tracking today, is the large number of hyper parameters required; this may harm robustness especially for real applications in unconstrained environments.

During the past few years, deep neural networks (DNNs) have become popular for their capability to learn rich features. Accordingly, new approaches with DNNs for tracking have also been proposed \cite{DBLP:journals/corr/GanGZC15,DBLP:journals/corr/KahouMM15,DBLP:journals/corr/OndruskaP16}. These methods take advantage of Recurrent Neural Networks (RNNs) to incorporate temporal information. Although some of these methods outperform conventional ones, computational requirements are high, resulting in very low frame rates and latency. Nevertheless, temporal information such as motion flow is crucial in object tracking, therefore cannot be discarded from a model without loss of performance. To address these issues, we present a new high speed tracking system, combining both appearance and temporal geometric information, while having a smaller number of hyper parameters. We achieve this by leveraging our newly designed Enhanced Siamese Neural Network (ESNN) architecture for similarity mapping: the ESNN is an extended Siamese neural network that combines appearance similarity with temporal geometric information and efficiently learns both visual and geometric features during end-to-end training. 


\section{Background} \label{sec:Background}
Although multiple object tracking plays a key role in computer vision, there exist few benchmarks  for pedestrian tracking, fewer than for object detection  \cite{Geiger2012CVPR,Dollar:2012:PDE:2197081.2197275,DBLP:journals/corr/LinMBHPRDZ14,ILSVRC15}. One reason is the difficulty in standardizing the evaluation protocol, a controversial topic this day \cite{DBLP:journals/corr/LuoZK14}; another reason may be high annotation cost. MOT16 \cite{DBLP:journals/corr/MilanL0RS16} and KITTI tracking benchmarks \cite{Geiger2012CVPR} provide well established evaluation protocols with good quality annotations, and are widely used by researchers. MOT16 consists of 14 different sequences and KITTI consists of 50 sequences. Whereas KITTI videos are taken with moving cameras (attached to a vehicle), MOT sequences are taken with both static and moving ones. Also, even though both datasets contain multiple objects types such as cars, cyclists, pedestrians, and motorbikes, KITTI evaluates only on cars and pedestrians and MOT16 evaluates only pedestrians. For fair comparison, MOT16 evaluation specifies additional information used by each submitted methods, for example, whether a method is online (no latency), and is using provided detection results. 

\begin{figure}
	\centering
	\includegraphics[height=3.8cm]{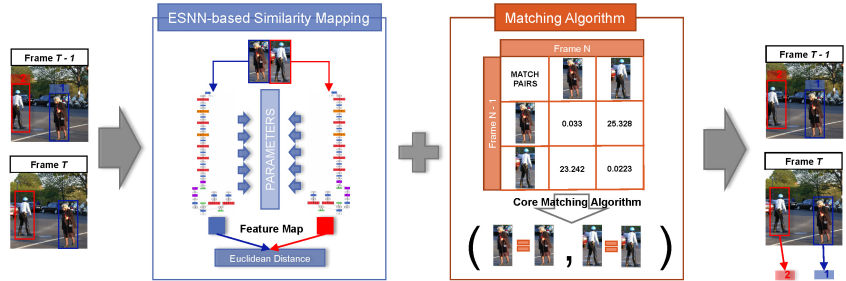}
	\caption{ESNN-based Multi-Object Tracking System}
	\label{fig:ESNNTrackingSystemArch}
\end{figure}

In this paper, we propose an online system based on provided detection results for two main reasons: first, we focus on visual tracking for ADAS and autonomous driving, and we believe reliable/low-latency tracking system is crucial. 
Secondly, since detection performance highly affects tracking quality and we want to focus our efforts on improving the tracking algorithm, we choose to use provided detection results for fair comparison. Fig.~\ref{fig:ESNNTrackingSystemArch} illustrates our tracking system based on ESNN. The system can be divided into two main steps: 1) ESNN-based Similarity Mapping and 2) Matching. A Siamese network, referred to as `Base Network', is pre-trained with visual information of objects. Then, ESNN takes Intersection-over-Union (IoU) and Area Ratio information from pairs of objects as additional features, and builds a combined similarity mapping with both geometric and pre-trained Siamese network features. After ESNN is fully trained and similarity scores are computed, the matching algorithm produces the final tracking results.

\section {Similarity Mapping} \label{sec:simMapping}

ESNN uses a Siamese network that consists of two identical sets of convolutional neural networks, where the weights of convolutional layers are shared in between. The network takes a pair of image patches, cropped from original frame, and then maps them to $L2$ space where the Euclidean distance between each output can be used as similarity score. The Base Network is built and trained first, then is extended to ESNN with geometric information.

\subsection{Base Network Architecture} \label{subsubsec:BaseArchitecture}
The base architecture of our Siamese neural network is described in Fig.~\ref{fig:BaseNetArch}. For each convolutional layer, {\it hyperbolic tangent} ({\it TanH}) is used as activation function, and the first fully connected layer is followed by {\it Rectified Linear Unit} ({\it ReLU}) \cite{Hinton_rectifiedlinear}. Kernel sizes for each convolutional and pooling layer are as follows: conv1(5x5), pool1(2x2), conv2(3x3), pool2(2x2), conv3(3x3), conv4(2x2), conv5(2x2), fc1(2048), fc2(1024), and feat(2) or feat(4). The feat(2) layer is fine-tuned with the new feat(4) layer to incorporate geometrical featurs in ESNN. For loss function, contrastive loss  $L_c$, proposed in \cite{Hadsell06dimensionalityreduction}, is used as follows:
\begin{figure}
	\centering
	\includegraphics[height=5.0cm]{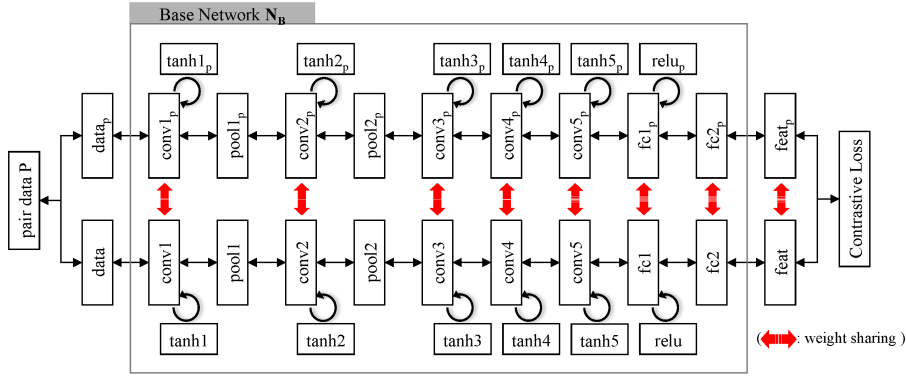}
	\caption{Architecture of Base Siamese Neural Network for Similarity Training}
	\label{fig:BaseNetArch}
\end{figure}
\noindent
\begin{gather}
E_n = \|F - F_p\|_2\\
L_c = \frac{1}{2N}\sum_{n=1}^{N} (y)E_n^2 + (1-y)\max(m-E_n, 0)^2
\end{gather}

where $E_n$ is Euclidean distance between the output features $F$ and $F_p$ of the Siamese neural network with input data pair $d$ and $d_p$, shown in Fig.~\ref{fig:BaseNetArch}. $y$ denotes label of the pair, where $y=1$ if ($d$, $d_p$) is a matching pair and $y=0$ otherwise. Finally, $m$ is a margin parameter that affects contribution of non-matching pairs to the loss $L_c$, and we choose $m=3$ as the best margin obtained by experiments. 

\subsection{ESNN Architecture} \label{subsubsec:ExArchitecture}
In extension of the Base Network architecture above, the ESNN takes additional layers that learn from IoU $D_{iou}$, and area variant of a pair of objects $D_{Arat}$. For a pair of object bounding boxes $b_i$ and $b_j$, appearing in frame $f_{t-1}$ and $f_t$, $D_{IoU}$ and $D_{Arat}$ are calculated as follows:
\noindent
\begin{align} \label{eq:IoUnArat}
[D_{IoU}, D_{Arat}](b_i, b_j) & = [\frac{area(b_i \cap b_j)}{area(b_i \cup b_j)}, \frac{min(area(b_i), area(b_j))}{max(area(b_i), area(b_j))}]
\end{align}

\begin{figure}
	\centering
	\includegraphics[height=5.3cm]{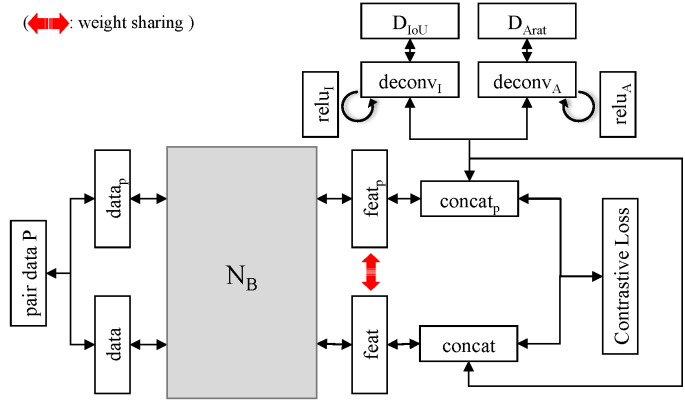}
	\caption{Architecture of Enhanced Siamese Neural Network}
	\label{fig:EsnnArchFig}
\end{figure}

Fig.~\ref{fig:EsnnArchFig} shows the extended architecture of our network. The additional layers up-sample input to the same dimension as the output of the Base Network $N_B$, $feat$ and $feat_p$. Layers in $N_B$ are locked during the first phase of training.

\subsection {Training} \label{subsec:Training}
\begin{figure}[htp]
	\centering
	\subfloat{{\includegraphics[scale=0.25]{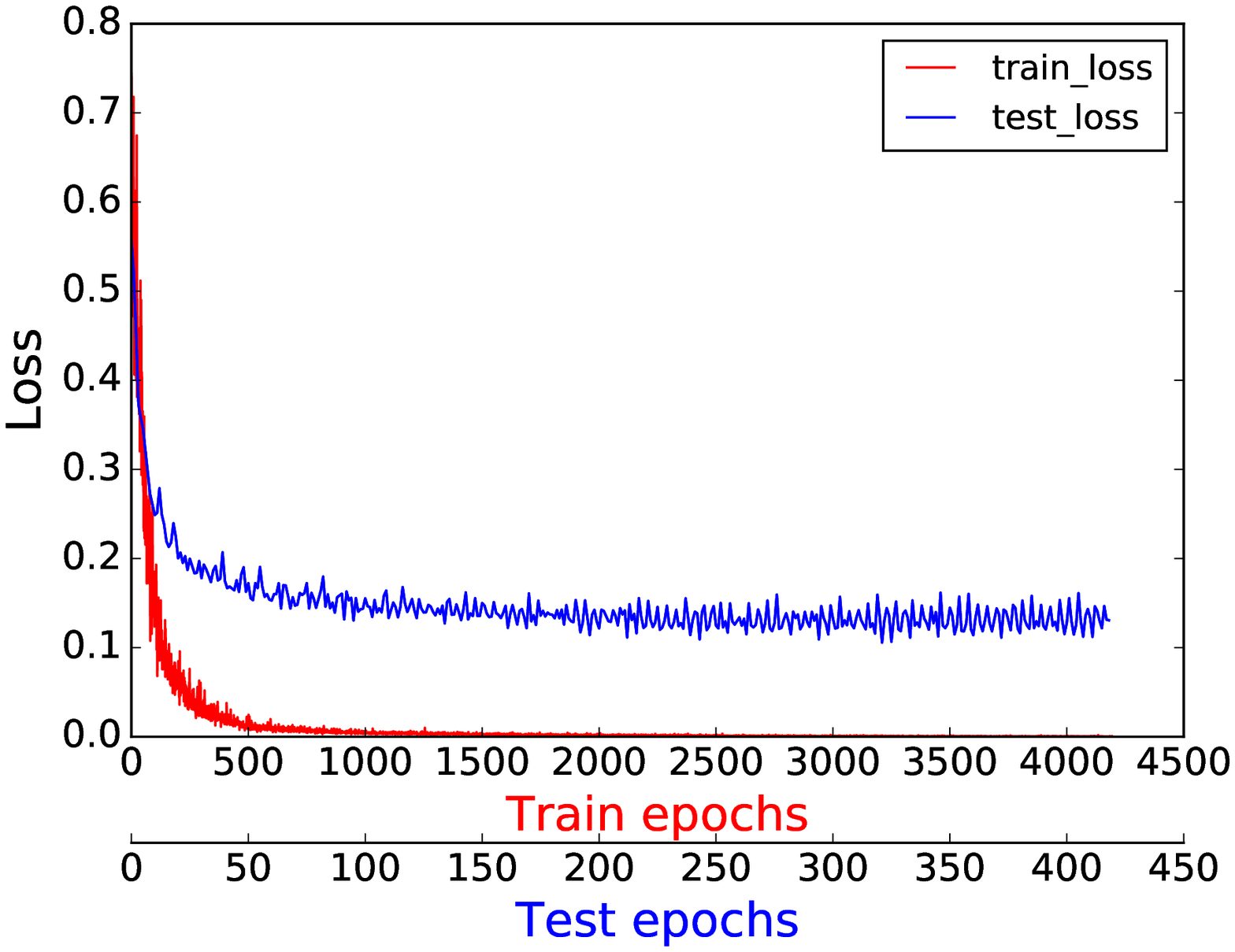} }}%
	\qquad
	\subfloat{{\includegraphics[scale=0.32]{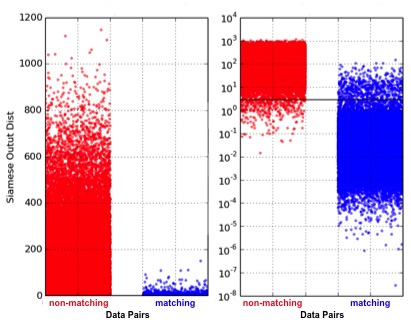} }}%
	\caption{Train/Test Loss of the Base Network ({\it left}) and Euclidean distance of Market-1501 test set with margin, $m=3$ (black horizental line) ({\it right})}
	\label{fig:smTrainPlot}%
\end{figure}

The Base Network is pre-trained on Market-1501 person re-identification dataset \cite{zheng2015scalable} first. With batch size of 128, learning rate starting from 0.01, and SGD (Stochastic Gradient Descent), our Siamese neural network converges well on pairs generated by Market-1501 dataset. Train and test losses of the training are shown in Fig.~\ref{fig:smTrainPlot} ({\it left}). {\it x-axis} represents the number of epochs in two different scales for each loss. On Market-1501 test set, the trained model achieves $precision=0.9854$, $recall=0.9774$, and $F_1 = 0.9814$.
In addition, Fig.~\ref{fig:smTrainPlot} ({\it right}) shows the Euclidean distance of the data pairs generated from the trained model on Market-1501 test set in logarithm scale ({\it y-axis}). With this pre-trained model, the network is then fine-tuned on MOT16 dataset. Results will be discussed at the end of this section along with ESNN training results.

\begin{figure}[htb]
	\captionsetup[subfloat]{farskip=2pt,captionskip=1pt}
	\centering
	\subfloat[Base Network Model]{\includegraphics[width=0.48\textwidth]{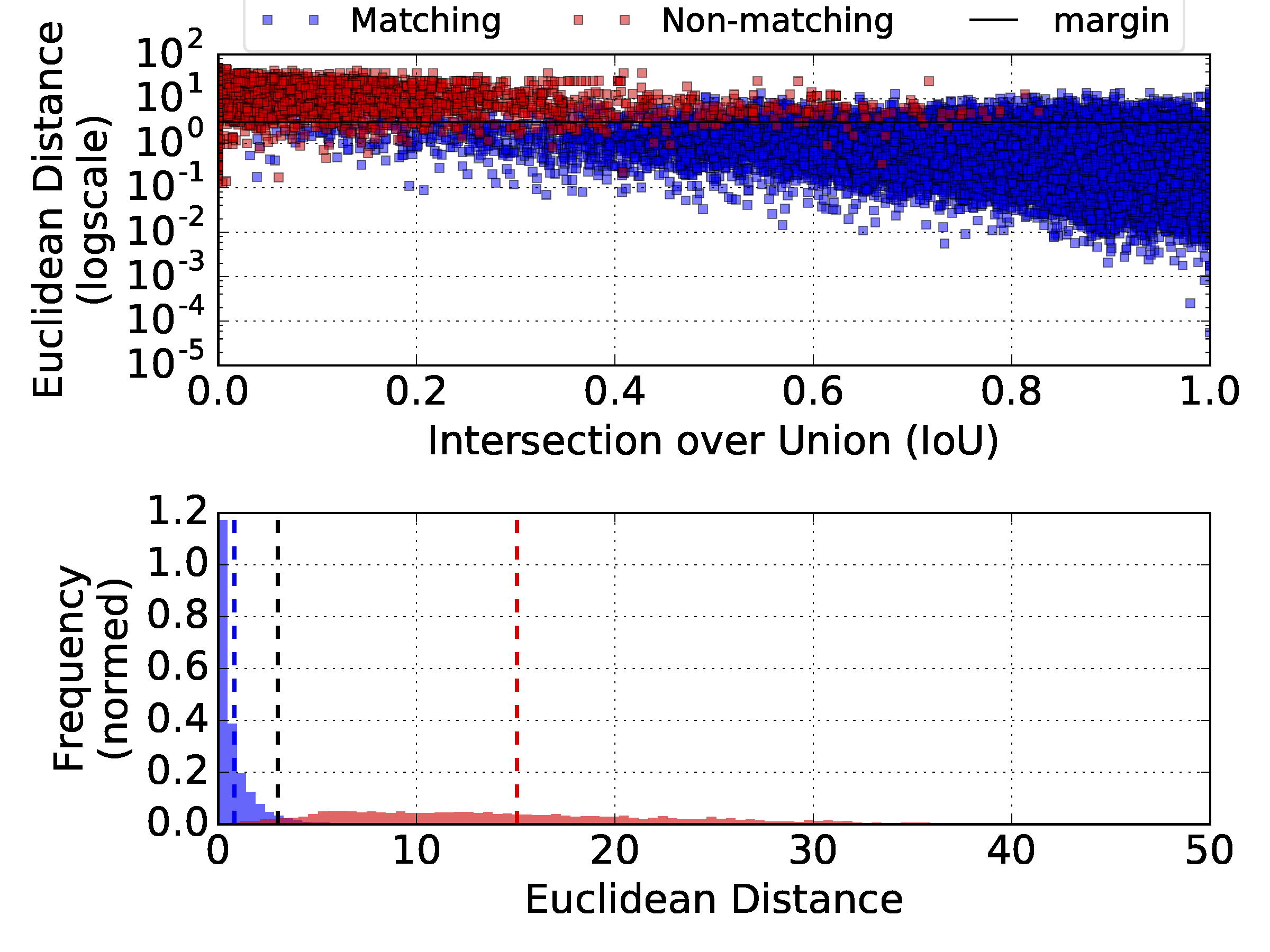}}\hfill
	\subfloat[ESNN Model]{\includegraphics[width=0.48\textwidth]{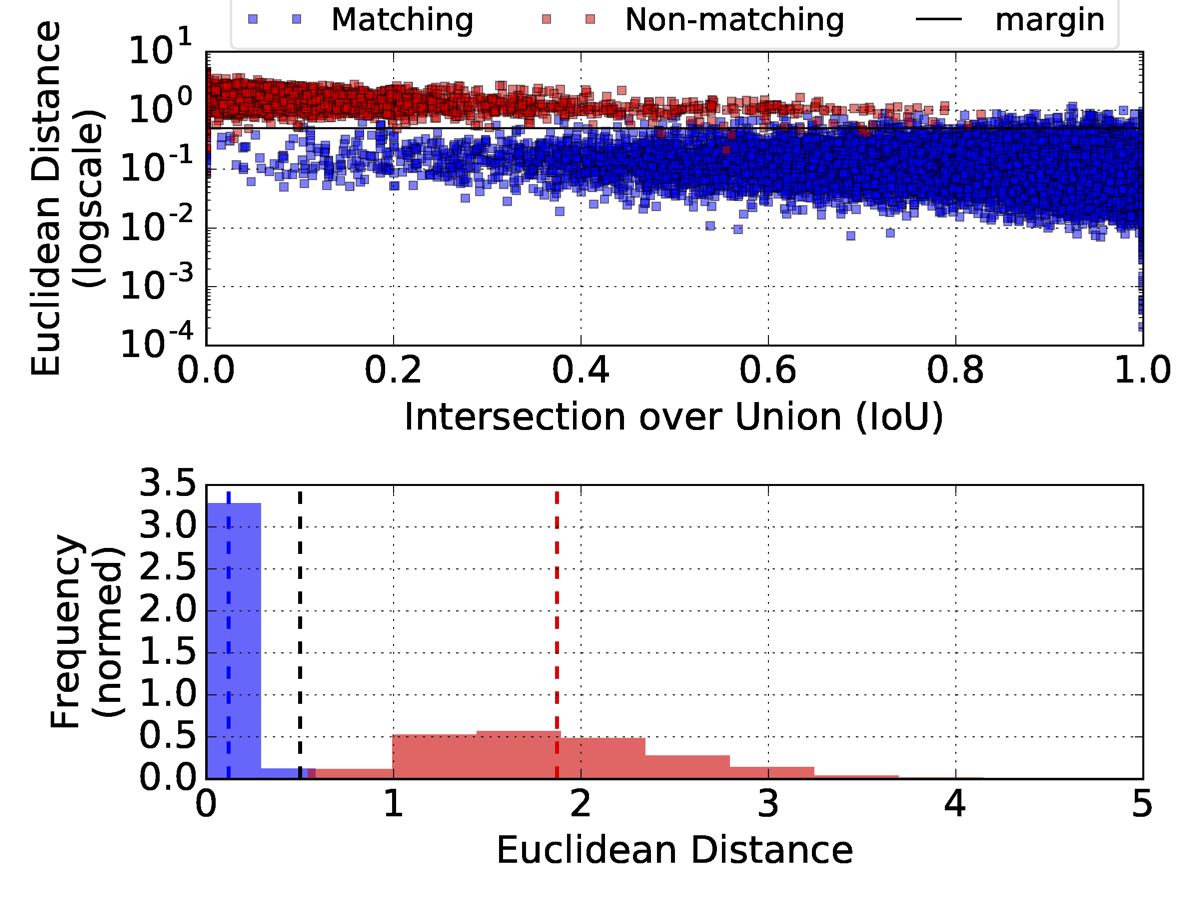}}
	\caption{Euclidean Distance on MOT16 Train Set}
	\label{fig:EDonMOT}
\end{figure}

To train ESNN, the pre-trained Base Network model parameters are transferred. In fine-tuning, layers in the Base Network are locked in the beginning, and unlocked in the final phase. Also, margin is set to $m=0.5$. Once the ESNN model is obtained, we analyze it on MOT16 train set, and compare the performance with results from the Base Network. Fig.~\ref{fig:EDonMOT} shows the Euclidean distance of MOT16 train set from the Base Network ({\it left}) and ESNN ({\it right}). On each figure, the plot on  top represents the Euclidean distance ({\it y-axis}) with IoU ({\it x-axis}) of the data. The bottom plot shows histogram of the Euclidean distance ({\it x-axis}) with normalized frequency ({\it y-axis}). The red points represent non-matching pairs,  blue points for matching pairs, and red and blue dashed lines represent mean distance of each group. Finally, the black dashed line represents the margin $m$. The Base Network model achieves $precision=0.9837$, $recall=0.9966$, and $F_1 = 0.9901$, and the ESNN model achieves {\it precision=0.9908}, {\it recall=0.9990}, and $F_1 = 0.9949$. As shown in Fig.~\ref{fig:EDonMOT}, the ESNN model outperforms the Base model. Note that, some of the misclassified non-matching pairs with $D_{IoU} < 0.05$ by the Base Network model are correctly classified by the ESNN model. It means the ESNN can handle object pairs spatially far apart but sharing similar features (e.g. two far-apart persons with similar clothing), better than the Base Network by utilizing IoU and area variant information.

\section {Matching Algorithm} \label{sec:matchingAlgorithm}
For the Base Network, a new score function is introduced by taking IoU and area variant in account, plus the score from Euclidean distance. For the ESNN, only Euclidean distance is used for scoring.
\begin{algorithm}[ht]
\small
	\caption{Matching Algorithm}\label{alg:matchingAlgo}
 
	\begin{algorithmic}[1]
		\Procedure{Match}{$P$, $f_n$}\Comment{Score matrix as input}
		\State $exID \gets$ \{\it {existing IDs whithin previous n frames}\}
		\For{$i$ in $reversed(sorted(P, \it score))$}\Comment{sort pairs with score}
		\State $(ID_{exist}, ID_{tgt})\gets P[i]$\Comment{pull candidate pair}
		\If{$ID_{tgt} \notin exID$}
		\State {continue}
		\EndIf
		\If{$ID_{tgt}.notAssigned()$}
		\If{$ID_{exist}.notAssigned()$}
		\State $Target[ID_{exist}] \gets ID_{tgt}$\Comment{new assignment}
		\State $ID_{exist}.setAssigned(True)$
		\ElsIf{ $newID_{exist} \gets FindBetterMatch()$}
		\State $ID_{exist}.setAssigned(False)$
		\State $Target[newID_{exist}] \gets ID_{tgt}$\Comment{switch assignment}
		\State $newID_{exist}.setAssigned(True)$
		\EndIf
		\EndIf
		\EndFor
		\For{$ID_{tgt} \in $ \{\it {leftover tgt IDs}\}}
		\State $exID.append(ID_{tgt}, f_{new})$\Comment{handle new IDs with frame info}
		\EndFor
		\EndProcedure
	\end{algorithmic}
\end{algorithm}
\subsection {Scoring} \label{subsec:scoring}

 Given detection boxes $B_{t-1} = \{b_1,\ldots,b_n\}$ at frame $t-1$, and $B_{t} = \{b_1,\ldots,b_m\}$ at frame $t$, new score function for a pair $S_{New}(b_i, b_j)$ where $b_i\in B_{t-1}$, $\forall i={1,\ldots,n}$, and $b_j\in B_{t}$, $\forall j={1,\ldots,k}$, can be derived as follows: 
\noindent
\begin{align}\label{eq:scorEq}
S_{New} & = S_{Dist} + S_{IoU}S_{Arat}
\end{align} where $S_{Dist}$ denotes the score derived from the Euclidean distance $D_{siam}(b_i, b_j)$, output of our  network for the pair $(b_i, b_j)$, $S_{IoU}$ denotes Intersection-over-Union of the pair, shifted by 1.0, and $S_{Arat}$ denotes the area ratio between them. To shorten notation, $S(b_i, b_j)$ is written as S in Eq.~\ref{eq:scorEq}. The exact functions of $S_{Dist}$, $S_{IoU}$, and $S_{Arat}$ are:
\noindent
\begin{align}
S_{Dist}(b_i, b_j) & = \alpha\log_{0.1}\{{max(\gamma, D_{siam}(b_i, b_j))}\}\\
S_{IoU}(b_i, b_j) & = 1.0 + \frac{area(b_i \cap b_j)}{area(b_i \cup b_j)}\\
S_{Arat}(b_i, b_j) & = e^{\frac{min(area(b_i), area(b_j))}{max(area(b_i), area(b_j))} - \delta}
\end{align}

where we choose $\alpha=0.8$, $\gamma=10^{-5}$, and $\delta=0.2$ as a bias term. Finally, $S_{New}$ is obtained for the Base Network model, and $S_{Dist}$ for the ESNN model.
\noindent
\begin{equation}
S_{Net} =
\begin{cases}
S_{New}, & \text{if}\ Net = N_B\\
S_{Dist}, & \text{otherwise}
\end{cases}
\end{equation}

\subsection {Matching} \label{subsec:matching}
As the second part of the tracking system, a simple yet efficient matching algorithm that takes the score matrix $S_{Net}$ as an input is derived as shown in Algorithm.~\ref{alg:matchingAlgo}. Only one hyper parameter is introduced by the algorithm, denoted by $f_n$, specifying how many frames the tracker looks back to generate pairs with the current frame. With $f_n$ and $S_{Net}$ map where data pair $P$ is the keys, the algorithm starts matching with the highest similarity score. It returns the best match solely based on the scores, and when there is a conflict, it tries once more to find a better match which can be replaced with the current match. After all possible pairs are examined and redundant pairs are filtered, new IDs are assigned to the leftover targets.

To provide a deeper insight on the advantages of this algorithm, we also employ a matcher based on the popular Hungarian algorithm and report the obtained results in the experimental section. One of the major differences between our proposed matching strategy and the Hungarian algorithm is computational complexity. In fact, while the former runs in linear time with the number of people in the scene, the Hungarian algorithm has a complexity of $O(n^3)$ and can become a significant performance bottleneck in crowded sequences.

\section {Evaluation} \label{sec:evaluation}
Our system is evaluated on MOT16 train and test set, as well as on the KITTI Object Tracking Evaluation 2012 database. 
The results on MOT16 test set is shown in Table~\ref{table:bcmkMOT16comparison}, along with other methods for comparison. Only the referencible methods that use provided detection results are shown, along with an indication whether the method is online or not. Table \ref{tab:kitti} reports the results on the KITTI database for the two evaluated classes, namely \textit{Car, Pedestrian}. Notice that no fine-tuning has been performed on the KITTI sequences, and the network has never seen objects from the \textit{Car} class during training. Nonetheless, the proposed algorithm achieves competitive performance, showing the good generalization capabilities of our architecture.

Even though an accurate comparison on speed is not quite possible due to lack of information on hardware specification where other benchmarks were conducted, the speed of our method is quite noticeable while achieving competitive performance. 

\setlength{\tabcolsep}{4pt}
\begin{table}
	\begin{center}
		\caption{Benchmark Results on MOT16 Test Dataset \cite{MOT16Web}}
		\label{table:bcmkMOT16comparison}
		\begin{tabular}{*{17}l}
			\hline\noalign{\smallskip}
			Method & Online & MOTA & MOTP & Hz & FAF & MT & ML & FP & FN & IDs & Frag\\
			\noalign{\smallskip}
			\hline
			\noalign{\smallskip}
			NMOT \cite{DBLP:journals/corr/Choi15} & No & 46.4 & 76.6 & \it 2.6 & 1.6 & 18.3\% & 41.4\% & 9,753 & 87,565 & 359 & 504\\			
			JMC \cite{conf/cvpr/TangAAS15} & No & 46.3 & 75.7 & \it 0.8 & 1.1 & 15.5\% & 39.7\% & 6,373 & 90,914 & 657 & 1,114\\
			MHT\_DAM \cite{kim_ICCV2015_MHTR} & No & 42.8 & 76.4 & \it 0.8 & 1.2 & 14.6\% & 49.0\% & 7,278 & 96.607 & 462 & 625\\
			\bf Ours & Yes & 35.3 & 75.2 & \it 7.9 & 0.9 & 7.4\% & 51.1\% & 5,592 & 110,778 & 1,598 & 5,153\\
			TBD \cite{10.1109/TPAMI.2013.185} & No & 33.7 & 76.5 & \it 1.3 & 1.0 & 7.2\% & 54.2\% & 5,804 & 112,587 & 2,418 & 2,252\\
			CEM \cite{Milan:2014:CEM} & No & 33.2 & 75.8 & \it 0.3 & 1.2 & 7.8\% & 54.4\% & 6,837 & 114,322 & 642 & 731\\
			DP\_NMS \cite{Pirsiavash:2011:GGA:2191740.2191761} & No & 32.2 & 76.4 & \it 212.6 & 0.2 & 5.4\% & 62.1\% & 1,123 & 121,579 & 972 & 944\\
			SMOT \cite{DicleICCV13} & No & 29.7 & 75.2 & \it 0.2 & 2.9 & 5.3\% & 47.7\% & 17,426 & 107,552 & 3,108 & 4,483\\
			JPDA\_m \cite{Rezatofighi:2015:ICCV} & No & 26.2 & 76.3 & \it 22.2 & 0.6 & 4.1\% & 67.5\% & 3,689 & 130,549 & 365 & 638\\
			\hline
		\end{tabular}
	\end{center}
\end{table}

Given the score matrix $S_{Net}$ provided by the siamese network, we compare the performance of the proposed matching algorithm to a baseline that uses the widely adopted Hungarian algorithm. The proposed matching approach is generally better than the Hungarian algorithm, who scores a MOTA of $27.7\%$. While a complete evaluation is omitted due to space constraints, it is worth noticing that besides resulting in a lower MOTA, the Hungarian algorithm is on average 1.91 times slower. In particular, while the execution time is substantially unchanged in some scenarios such as MOT16-05 (1.03 times slower), the Hungarian's $O(n^3)$ scalability is especially clear when dealing with the most crowded scenes, e.g. MOT16-04 (2.69 times slower).

\setlength{\tabcolsep}{4pt}
\begin{table}[t]
	\begin{center}
		\caption{Results on KITTI MOT Dataset using public(top) \& private(bottom) detections}
		\label{tab:kitti}
		\begin{tabular}{*{15}l}
			\hline\noalign{\smallskip}
			Name& MOTA & MOTP & MOTAL & Hz & Rcll & Prcn & FAR & MT & PT & ML & FP & FN & IDs & FM \\
			\noalign{\smallskip}
			\hline
			\noalign{\smallskip}
			Car        & 65.97 & 79.31 & 66.43 & 7.52 & 76.47 & 91.45 & 24.45 & 44.21 & 45.12 & 10.67 & 2723 &  8963 & 161 & 969\\
			Pedestrian & 33.69 & 70.46 & 34.42 & 11.81 & 44.22 & 82.13 & 20.19 & 10.31 & 52.23 & 37.45 & 2246 & 13024 & 172 & 1212\\
			\hline
		\end{tabular}
	\end{center}
\end{table}
\begin{table}[t]
	\begin{center}
		\caption{Results on KITTI MOT Dataset using private detections}
		\label{tab:kitti_private}
		\begin{tabular}{*{15}l}
			\hline\noalign{\smallskip}
			Name& MOTA & MOTP & MOTAL & Hz & Rcll & Prcn & FAR & MT & PT & ML & FP & FN & IDs & FM \\
			\noalign{\smallskip}
			\hline
			\noalign{\smallskip}
			Car        & 70.78 & 80.38 & 71.25 & 7.52 & 79.18 & 92.71 & 20.72 & 51.68 & 40.55 & 7.77 & 2305 &  7701 & 169 & 938\\
			Pedestrian & 37.04 & 71.13 & 37.90 & 11.81 & 46.56 & 84.53 & 17.88 & 14.09 & 56.36 & 29.55 & 1989 & 12473 & 202 & 1270\\
			\hline
		\end{tabular}
	\end{center}
\end{table}

		

\section {Conclusion} \label{sec:conclusion}
In this paper, we proposed a new approach for multiple object tracking system that takes advantage of deep Siamese neural network to model similarity mapping, followed by an efficient matching algorithm. We showed the capability of our Enhanced Siamese neural network, that can fuse appearance features with geometric information such as IoU and area variant of objects, resulting in better performance while keeping no latency. Evaluation results show that using Siamese neural network has significant potential for building similarity matrices for multiple object tracking. 

\medskip

\small
\clearpage
\bibliographystyle{splncs}
\bibliography{SM_ESNN_MOT}

\begin{thebibliography}{10}

\bibitem{DBLP:journals/corr/abs-1303-4803}
Li, X., Hu, W., Shen, C., Zhang, Z., Dick, A.R., van~den Hengel, A.:
\newblock A survey of appearance models in visual object tracking.
\newblock CoRR \textbf{abs/1303.4803} (2013)

\bibitem{Hu:2004:SVS:2220414.2220805}
Hu, W., Tan, T., Wang, L., Maybank, S.:
\newblock A survey on visual surveillance of object motion and behaviors.
\newblock Trans. Sys. Man Cyber Part C \textbf{34}(3) (August 2004)  334--352

\bibitem{Geiger2012CVPR}
Geiger, A., Lenz, P., Urtasun, R.:
\newblock Are we ready for autonomous driving? the kitti vision benchmark
  suite.
\newblock In: Conference on Computer Vision and Pattern Recognition (CVPR).
  (2012)

\bibitem{WuLimYang13}
Wu, Y., Lim, J., Yang, M.H.:
\newblock Online object tracking: A benchmark.
\newblock In: IEEE Conference on Computer Vision and Pattern Recognition
  (CVPR). (2013)

\bibitem{DBLP:journals/corr/GanGZC15}
Gan, Q., Guo, Q., Zhang, Z., Cho, K.:
\newblock First step toward model-free, anonymous object tracking with
  recurrent neural networks.
\newblock CoRR \textbf{abs/1511.06425} (2015)

\bibitem{DBLP:journals/corr/KahouMM15}
Kahou, S.E., Michalski, V., Memisevic, R.:
\newblock {RATM:} recurrent attentive tracking model.
\newblock CoRR \textbf{abs/1510.08660} (2015)

\bibitem{DBLP:journals/corr/OndruskaP16}
Ondruska, P., Posner, I.:
\newblock Deep tracking: Seeing beyond seeing using recurrent neural networks.
\newblock CoRR \textbf{abs/1602.00991} (2016)

\bibitem{Dollar:2012:PDE:2197081.2197275}
Dollar, P., Wojek, C., Schiele, B., Perona, P.:
\newblock Pedestrian detection: An evaluation of the state of the art.
\newblock IEEE Trans. Pattern Anal. Mach. Intell. \textbf{34}(4) (April 2012)
  743--761

\bibitem{DBLP:journals/corr/LinMBHPRDZ14}
Lin, T., Maire, M., Belongie, S.J., Bourdev, L.D., Girshick, R.B., Hays, J.,
  Perona, P., Ramanan, D., Doll{\'{a}}r, P., Zitnick, C.L.:
\newblock Microsoft {COCO:} common objects in context.
\newblock CoRR \textbf{abs/1405.0312} (2014)

\bibitem{ILSVRC15}
Russakovsky, O., Deng, J., Su, H., Krause, J., Satheesh, S., Ma, S., Huang, Z.,
  Karpathy, A., Khosla, A., Bernstein, M., Berg, A.C., Fei-Fei, L.:
\newblock {ImageNet Large Scale Visual Recognition Challenge}.
\newblock International Journal of Computer Vision (IJCV) \textbf{115}(3)
  (2015)  211--252

\bibitem{DBLP:journals/corr/LuoZK14}
Luo, W., Zhao, X., Kim, T.:
\newblock Multiple object tracking: {A} review.
\newblock CoRR \textbf{abs/1409.7618} (2014)

\bibitem{DBLP:journals/corr/MilanL0RS16}
Milan, A., Leal{-}Taix{\'{e}}, L., Reid, I.D., Roth, S., Schindler, K.:
\newblock {MOT16:} {A} benchmark for multi-object tracking.
\newblock CoRR \textbf{abs/1603.00831} (2016)

\bibitem{Hinton_rectifiedlinear}
Hinton, G.E.:
\newblock Rectified linear units improve restricted boltzmann machines vinod
  nair

\bibitem{Hadsell06dimensionalityreduction}
Hadsell, R., Chopra, S., Lecun, Y.:
\newblock Dimensionality reduction by learning an invariant mapping.
\newblock In: In Proc. Computer Vision and Pattern Recognition Conference
  (CVPR’06), IEEE Press (2006)

\bibitem{zheng2015scalable}
Zheng, L., Shen, L., Tian, L., Wang, S., Wang, J., Tian, Q.:
\newblock Scalable person re-identification: A benchmark.
\newblock In: Computer Vision, IEEE International Conference on. (2015)

\bibitem{MOT16Web}
:
\newblock Multiple object tracking benchmark.
\newblock \url{https://motchallenge.net/results/MOT16/}

\bibitem{DBLP:journals/corr/Choi15}
Choi, W.:
\newblock Near-online multi-target tracking with aggregated local flow
  descriptor.
\newblock CoRR \textbf{abs/1504.02340} (2015)

\bibitem{conf/cvpr/TangAAS15}
Tang, S., Andres, B., Andriluka, M., Schiele, B.:
\newblock Subgraph decomposition for multi-target tracking.
\newblock In: CVPR, IEEE Computer Society (2015)  5033--5041

\bibitem{kim_ICCV2015_MHTR}
Kim, C., Li, F., Ciptadi, A., Rehg, J.M.:
\newblock Multiple hypothesis tracking revisited.
\newblock In: Computer Vision (ICCV), IEEE International Conference on, IEEE
  (December 2015)

\bibitem{10.1109/TPAMI.2013.185}
Stiller, C., Urtasun, R., Wojek, C., Lauer, M., Geiger, A.:
\newblock 3d traffic scene understanding from movable platforms.
\newblock IEEE Transactions on Pattern Analysis and Machine Intelligence
  \textbf{36}(5) (2014)  1--1

\bibitem{Milan:2014:CEM}
Milan, A., Roth, S., Schindler, K.:
\newblock Continuous energy minimization for multitarget tracking.
\newblock IEEE TPAMI \textbf{36}(1) (2014)  58--72

\bibitem{Pirsiavash:2011:GGA:2191740.2191761}
Pirsiavash, H., Ramanan, D., Fowlkes, C.C.:
\newblock Globally-optimal greedy algorithms for tracking a variable number of
  objects.
\newblock In: Proceedings of the 2011 IEEE Conference on Computer Vision and
  Pattern Recognition. CVPR '11, Washington, DC, USA, IEEE Computer Society
  (2011)  1201--1208

\bibitem{DicleICCV13}
Dicle, C., Sznaier, M., Camps, O.:
\newblock The way they move: Tracking targets with similar appearance.
\newblock In: ICCV. (2013)

\bibitem{Rezatofighi:2015:ICCV}
Rezatofighi, S.H., Milan, A., Zhang, Z., Shi, Q., Dick, A., Reid, I.:
\newblock Joint probabilistic data association revisited.
\newblock In: ICCV. (2015)

\end{thebibliography}

\end{document}